# Comparative Study of Canvas- and Coordinate-Based Geometry-Conditioned Neural Networks for Real-Time Prediction of Indoor Environments with Varying Room Dimensions

Shaofan Wang, Philipp Geyer
Sustainable Building Systems Group, Leibniz University Hannover, Herrenhäuser Str. 8, 30419 Hannover, Germany

**Abstract.** Computational fluid dynamics (CFD) is computationally expensive for iterative building design. Deep learning surrogates offer faster predictions, but generalization across varying room dimensions remains challenging. This study compares two geometry-conditioned neural network surrogate models for rapid indoor airflow and temperature field prediction under varying room sizes. During the architectural design stage, this method is relevant for real-time assessment of whether conditioned supply air will effectively provide fresh and cool air throughout rooms with varying heights and lengths. The canvas-based neural network (CaNN) explicitly encodes room size by preserving different occupied regions within a fixed canvas, whereas the coordinate-based neural network (CoNN) uses normalized coordinate channels. Both employ U-Net autoencoders to separately predict airflow and temperature fields. Evaluation uses simplified rooms with varying heights and lengths as a controlled benchmark. The results show that CaNN achieves higher agreement with CFD results than CoNN, demonstrating its potential to support performance-driven building design.

## 1. Introduction

Indoor environmental quality is critical in occupant health, thermal comfort, and productivity [1], as people spend the majority of their time indoors, maintaining a healthy and energy-efficient indoor environment has become an increasingly important objective in modern building design and operation [2,3]. In this context, indoor environment prediction therefore plays an important role in building design, ventilation assessment, thermal comfort analysis, indoor air quality evaluation, and HVAC system optimization [4]. With the increasing emphasis on energy-efficient and healthy buildings, rapid and accurate evaluation of indoor airflow and temperature performance during the early stages of building design has become increasingly important for supporting informed design decisions [5–7]. However, detailed airflow analysis is often performed only at later stages of the design process, when design flexibility has already diminished and modifications become considerably more expensive [8]. Therefore, there is a growing demand for rapid and high-fidelity prediction methods capable of providing detailed indoor airflow information during early-stage design exploration.

Computational fluid dynamics (CFD) has become one of the most widely used approaches for indoor environmental analysis for decades because it provides detailed spatial distributions of airflow velocity, temperature, and contaminant transport [9]. By numerically solving the governing fluid equations, CFD enables high-resolution prediction of complex indoor flow phenomena, including jet development, recirculation structures, stratified airflow, and thermal plume interactions [10]. Compared with simplified multi-zone airflow models based on well-mixed assumptions, CFD offers significantly higher physical fidelity and stronger capability for local environmental analysis [11]. However, despite its high accuracy, CFD simulation is computationally expensive and time-consuming, especially for large parametric studies, geometry optimization, and real-time control applications [12]. The computational cost becomes even more significant when huge design space and parameters need to be evaluated repeatedly. This limitation restricts the applicability of CFD in early-stage design exploration, rapid decision-making, and real-time building operation [13]. Consequently, to alleviate the computational burden of conventional CFD, several reduced-cost numerical approaches have been proposed. For example, coarse-grid [14] CFD methods demonstrated the potential for significantly accelerated prediction, and Fast fluid dynamics (FFD) [15] further improved computational efficiency and achieved substantial reductions in simulation time compared with conventional CFD approaches. Nevertheless, the simplified numerical schemes and reduced spatial resolution employed in CFD modeling may compromise prediction accuracy, especially in regions involving strong gradients and complex flow structures [16,17]. From another perspective, GPU-accelerated lattice Boltzmann methods (LBM) have recently shown promising capability for indoor airflow and temperature prediction while maintaining reasonable agreement with traditional CFD approaches [18]. Despite these advancements, the computational cost of such physics-based methods remains prohibitively high for real-time prediction and large-scale design iteration tasks. In addition to the limitations mentioned above, CFD simulations require complex model setup procedures, including geometry preparation, mesh generation, boundary condition specification, and numerical parameter selection, all of which demand substantial expertise in fluid mechanics and numerical methods. Such requirements limit the accessibility of CFD tolls for architects and building designers, who often lack specialized CFD knowledge, thereby potentially leading to inaccurate or unreliable simulation results when improperly configured [19].

Therefore, there remains a strong need for surrogate modeling approaches capable of providing rapid, user-friendly and high-fidelity indoor airflow prediction with substantially reduced computational cost and improved accessibility for design-stage applications [20]. Recent developments in machine learning and reduced order modeling (ROM) have provided new opportunities for the development of data-driven surrogate models for CFD prediction [21,22]. Unlike conventional physics-based numerical approaches, surrogate models aim to learn the mapping relationship between flow conditions and corresponding flow fields. Existing surrogate modeling approaches are generally categorized along a spectrum ranging from relatively interpretable ROM based on linear modal decomposition to highly nonlinear deep learning frameworks capable of learning complex flow representations directly from data [23–26] . Early surrogate modeling approaches for indoor airflow prediction mainly relied on reduced-order modeling techniques. Kluftødegård and Chaudhuri [27] proposed a proper orthogonal decomposition (POD)-based interpolation framework using CFD snapshots to reconstruct displacement ventilation flow fields with substantially reduced computational cost. With recent development in neural networks (NN) have emerged as a promising alternative for nonlinear CFD surrogate modeling. Zhou and Ooka [28,29] systematically investigated deep neural network architectures and data pre-processing strategies for indoor airflow prediction with different boundary parameters using CFD-generated datasets. Their studies demonstrated that convolutional neural networks (CNN) and appropriate data pre-processing methods significantly improved the prediction accuracy of non-isothermal indoor airflow distributions while achieving substantially faster prediction speeds than conventional CFD simulations. Besides, Wang et al. [30] developed a deep autoencoder framework for indoor airflow prediction and compared its performance with POD-based reduced-order modeling under varying inlet velocities. Their results demonstrated that deep autoencoder frameworks exhibited stronger nonlinear representation capability and improved reconstruction performance for complex indoor airflow fields. Building upon these advances in nonlinear representation learning, subsequent studies further explored more complex neural network frameworks for rapid CFD surrogate prediction. Afterwards, some studies developed surrogate model of real-time indoor environment prediction for different inlet positions based on CNN and U-Net, which demonstrates that these models are able to reduce computational cost while preserving the major airflow patterns [31,32]. Zhang et al. [33] proposed a CNN-based surrogate framework for rapid indoor airflow prediction and introduced a geometry representation strategy to incorporate room geometry information into CNN inputs. Compared with conventional ANN-based and GNN-based surrogate models, the proposed CNN framework demonstrated superior capability in capturing local spatial airflow features, significantly improving prediction accuracy while reducing computational time to approximately 0.05% of that required for conventional CFD simulations. Liu et al. [34] proposed a composite ANN-VAE framework for rapid reconstruction and prediction of temperature and velocity fields for data center. The proposed model achieved high-accuracy flow field reconstruction while accelerating prediction speed by approximately 380,000 times compared with conventional CFD simulations. More recently, surrogate modeling approaches have also been extended toward high-resolution three-dimensional (3D) airflow prediction involving varying room layouts and inlet configurations. Zheng and Xu [35] developed a cascaded ANN framework for predicting 3D indoor airflow velocity and temperature distributions under varying air supply conditions and room layouts. By incorporating air inlet configurations and occupant locations as 3D inputs, the proposed model demonstrated strong generalization capability for unseen configurations while achieving prediction speeds suitable for real-time HVAC control applications. In addition to improving prediction accuracy and computational efficiency, recent studies have also focused on enhancing the physical consistency of data-driven airflow reconstruction models. Wei and Ooka [36] introduced a physics-informed neural network (PINN) framework for reconstructing indoor airflow fields directly from measurement data without requiring explicit inlet boundary conditions or large CFD databases. Compared with conventional ANN approaches, the proposed PINN demonstrated improved physical consistency and stronger robustness under sparse measurement conditions. Moreover, the framework reduced computational cost while avoiding inaccuracies associated with CFD boundary condition specification. To further improve interpretability, flexibility and reusability, Wang et al. [37] proposed a component-based machine learning (CBML) framework for indoor airflow and temperature prediction by decomposing complex indoor configurations into reusable flow components and aggregating their latent representations. The results demonstrated that latent feature aggregation could effectively capture flow interactions between multiple components while significantly reducing the computational cost and enabling reuse of per-trained models.

Despite significant progress, most existing surrogate models primarily focused on parametric interpolation problems within fixed computational domains. In many studies, the room geometry, mesh topology, and spatial dimensions remained unchanged, while only boundary conditions such as inlet velocity or temperature were varied. Under such settings, the learned mapping relationship mainly represents parameter-conditioned flow prediction rather than geometry-conditioned flow prediction. During the architectural design stage, room dimensions are frequently modified to satisfy functional, aesthetic, and energy-efficiency requirements. Such geometric changes substantially change airflow structures and temperature distributions affecting whether conditioned supply air is able to effectively reach the occupied zone and maintain comfortable indoor conditions or returning before reaching areas close to façade leading to discomfort in these areas. To ensure satisfactory ventilation performance, architects and HVAC engineers need to evaluate a large number of geometric design alternatives in early phases; due to computational load, modelling effort, and often missing expertise in CFD modeling, this evaluation often renders unfeasible based on traditional physical CFD in standard design processes. Consequently, they rely on gut feeling leading to significant uncertainty and a significant risk of design errors and discomfort in the realized building.

This use case highlights the need for surrogate models capable of accurately predicting airflow and thermal fields under varying geometrical configurations in real time. While systematic investigation of geometric generalization capability for indoor airflow surrogate modeling remains largely unexplored in the built environment field, related studies have begun to emerge in other CFD domains. These studies provide valuable insights into geometry representation and geometric generalization for surrogate modeling. Guo et al. [38] proposed a CNN-based surrogate framework for steady flow prediction around arbitrary geometries in aerodynamic applications and introduced signed distance functions (SDF) as an implicit geometry representation strategy for convolutional neural networks. Building upon this framework, Ribeiro et al. [39] further developed the DeepCFD model in aerodynamic field using U-Net architectures with skip connections to improve the reconstruction accuracy of coupled velocity and pressure fields for steady flow prediction. However, these studies primarily focused on external flow prediction problems involving varying obstacle geometries embedded within fixed computational domains. Under such settings, the global computational domain, inlet/outlet configurations, and overall spatial topology remained unchanged, while only internal object shapes varied. In contrast, indoor airflow prediction under varying room geometries represents a fundamentally different problem, where the computational domain itself changes with room dimensions and layouts. Such geometry variations directly modify the spatial support of the flow field and introduce substantially greater challenges for geometry-conditioned surrogate modeling and spatial generalization. This challenge arises because conventional CNN framework requires input tensors with consistent spatial dimensions for batch training, primarily for computational efficiency and spatial alignment in batch normalization and skip connections [40,41]. Domains of varying room sizes therefore require appropriate sampling or mapping to satisfy this requirement while preserving physical scale information. While SDFs encode geometry through signed distances to boundaries, they do not specify how computational domains of different room sizes and their associated flow and temperature fields are sampled or mapped into fixed-size tensors for CNN training. This gap makes it difficult to directly apply CNN backbones to variable-size indoor domains.

Motivated by the aforementioned research gap, this paper proposes two methods to create pre-trained reusable prediction models for changing room sizes: Canvas-based (CaNN) and Coordinate-based (CoNN) geometry-conditioned neural networks for rapid prediction of indoor airflow speed and temperature fields across parametrically varying room geometries. The main contribution of this study is the development and comparative investigation of two spatial representation strategies that map indoor flow domains of varying physical dimensions and their corresponding flow and temperature fields into fixed-size input-output tensors while retaining physical geometric information for CNN-based surrogate modeling. CaNN explicitly embeds solid-void-based computational domains into a fixed-grid-size spatial canvas, whereas CoNN implicitly represents geometry through only globally normalized spatial coordinates with a fixed number of grid points and changing grid size in real world dimensions. A rectangular indoor environment with varying room heights and lengths under identical boundary conditions is employed to evaluate the geometric generalization capability of both approaches. The prediction accuracy, spatial error distributions, and generalization performance of CaNN and CoNN are systematically compared to investigate the influence of geometry representations on indoor airflow surrogate modeling.

The rest of paper is presented as follows: First, the CFD simulation setup and parametrically varying indoor room configurations are introduced, together with the dataset generation procedure for airflow speed and temperature fields. Next, the methodology section explains the proposed geometry-conditioned surrogate modeling frameworks of CaNN and CoNN, with particular emphasis on their geometry representations. Then, the section of results and discussion systematically compares the prediction accuracy, spatial error distributions, and geometric generalization capability of both approaches under varying room dimensions. Finally, the main conclusions are summarized, and future research directions for geometry-conditioned CFD surrogate modeling are outlined.

## 2. Methodology

This part explains the methods of CaNN and CoNN including benchmark case, CFD simulation, dataset generation, geometry representation and NN structure. Specifically, section 2.1 introduces benchmark description and CFD model set up, as well as dataset generation for training, validating and testing. Then, in section 2.2, geometry representations and NN structure are provided.

### 2.1 CFD simulation and dataset construction

To construct a geometry-diverse CFD database for surrogate model development, a parametric workbench was established by systematically varying the room dimensions. The room geometry was parameterized using two independent variables: room length (L) and room height (H). Fig. 1 shows that the room length was varied from 3 m to 9 m with an increment of 0.5 m, while the room height ranged from 3 m to 5 m with the same increment. This parametrization generated a total of 65 room configurations covering a broad range of room aspect ratios and enclosure sizes. To isolate the influence of geometry on airflow patterns, all ventilation and thermal conditions were kept unchanged across all cases. The room was equipped with one inlet and one outlet, both with an opening size of 0.1 m. The inlet was positioned at the upper-left wall, whereas the outlet was located at the lower-right wall, forming a diagonal ventilation layout. The inlet velocity was fixed at 1 m/s, and the supply air temperature was maintained at 15 °C for all simulations. To represent summer cooling conditions, two windows were placed on the left and right walls, respectively. Unlike the ventilation openings, the window dimensions varied with room height. Specifically, each window occupies the entire wall surface except for the regions reserved for the ventilation openings, leaving 0.1 m clearance at both the top and bottom of the wall. Considering that the window was equipped with a shading device, a constant heat flux of 80 W/m² was imposed on both window surfaces to represent external thermal loads and gains by heat transmission [42]. Simple room geometries and fixed ventilation conditions were deliberately adopted to provide a controlled benchmark for assessing how geometry representation affects prediction accuracy and generalization across room sizes, while limiting the confounding effects of complex layouts and varying boundary conditions.

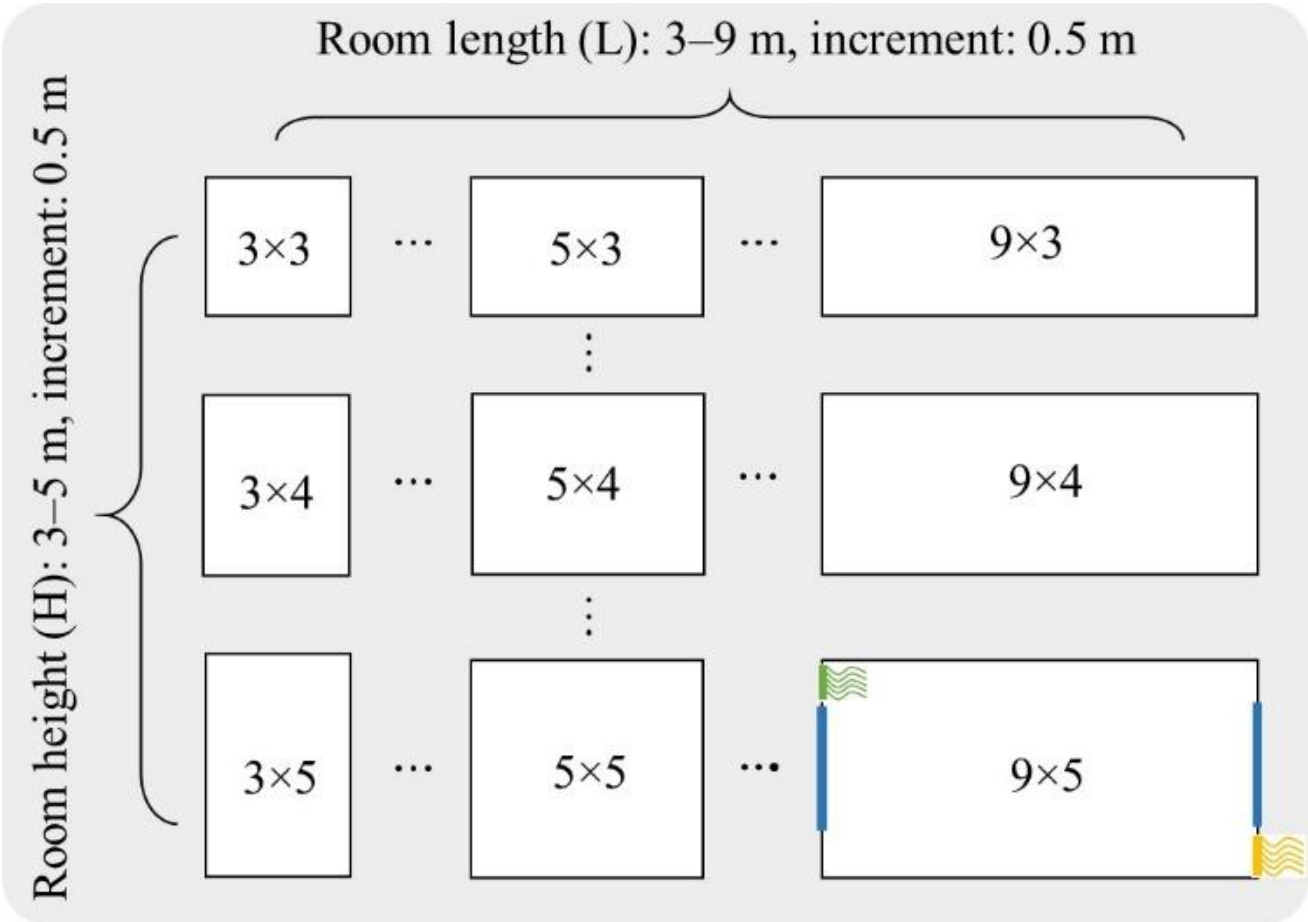

Fig. 1. Representative room geometries. The room length L varies from 3 m to 9 m and the room height H varies from 3 m to 5 m with increments of 0.5 m. All room configurations maintain identical inlet and outlet boundary conditions.

To generate the dataset of reference airflow and temperature fields for surrogate model training and evaluation, CFD simulations were employed by using OpenFOAM software. The airflow and heat transfer inside the room are governed by steady Reynolds-Averaged Navier-Stokes (RANS) framework involving the continuity, momentum, and energy equations, with turbulence model provided by the RNG k-epsilon. Detailed formulations of the governing equations are found in standard CFD literature and the OpenFOAM documentation [43]. Finite volume method (FVM) was employed to discretize the governing equations over computational domain [44]. Thermal buoyancy effects were incorporated using the Boussinesq approximation, in which density variation was considered only in the buoyancy term while all other fluid properties remained constant [45]. The airflow was supplied through a velocity inlet with a constant velocity of 1 m/s and a supply air temperature of 15 °C. The exhaust opening was defined as a pressure outlet. All solid surfaces were treated as no-slip walls. Besides, the windows were applied as heat flux wall with 80 W/m$^2$.

Structured meshes were generated for all room configurations using a consistent meshing strategy to ensure comparable spatial resolution across different geometries. To improve the prediction of near-wall transport and local flow development, mesh refinement was applied along solid boundaries and around the supply and exhaust openings, where strong velocity and temperature gradients were expected. In addition, near-wall functions were employed to model the turbulent boundary layer behavior, reducing the mesh resolution requirements close to solid surfaces while maintaining computational efficiency and prediction accuracy. To ensure the reliability of simulation outputs, mesh independence study was conducted. For each selected case, multiple mesh densities were tested by progressively reducing the cell size while keeping the refinement strategy unchanged. Since the room dimensions varied across cases, mesh resolution was characterized by the number of cells per unit area rather than the total cell count. Four mesh densities of approximately 200, 450, 700, and 950 cells/m² were evaluated under identical boundary conditions. The relative differences in area-averaged velocity and temperature between successive mesh densities were approximately 5%, 2.1%, and 1.5%, respectively. Based on the limited changes observed upon refinement from 700 to 950 cells/m², a minimum mesh density of approximately 700 cells/m² was adopted for all room geometries to balance computational accuracy and efficiency. This mesh generation approach was subsequently adopted for all simulations. Besides, before generation of dataset, the CFD model was validated against experimental data reported in [46], following the procedure described in previous study [37]. The comparison yielded relative error lower than 5% for u-velocity and temperature profiles at mid-plane position of the room in the validation case, which demonstrated satisfactory agreement with experiment and confirmed the reliability of CFD simulation. In addition, overall mass and energy balances were checked by accounting for inlet and outlet fluxes and window heat transfer, where applicable. Simulations were considered converged when the residuals of all solved variables satisfied the prescribed convergence criteria. The resulting velocity and temperature distributions were extracted as ground-truth data for surrogate model training and testing.

After generating the CFD database from the parametric workbench, the dataset was manually partitioned into training, validation, and test dataset based on prior physical knowledge of indoor airflow behavior, as illustrated in Fig. 2. Preliminary analysis of the simulated flow fields reveal that the generated cases could be categorized into two distinct airflow regimes depending on room dimensions. For rooms with relatively short lengths, the supply jet is able to reach the right wall and formed a dominant circulation occupying most of the room volume. Under this condition, a single large-scale recirculation structure governs the overall airflow pattern. As the room length increased, the airflow pattern changes from a single room-scale circulation to a two-circulation structure. For shorter rooms, the supply jet is sufficiently strong to drive a dominant circulation spanning the entire room. Beyond a critical room length, however, the jet momentum is not able to sustain circulation near the far wall. As the room becomes longer, the supply jet gradually loses momentum before reaching the far end of the room. Consequently, a strong clockwise recirculation remains in the left part of the room, while a weaker anticlockwise circulation emerges in the right region due to buoyancy effects associated with the heated window surfaces. This shift from a single-circulation to a double-circulation topology is regarded as a flow-pattern transition. The critical room length at which this transition occurred is found to depend on the room heights. Cases located near these pattern-transition cases exhibit stronger flow dynamic and larger sensitivity to geometric variation. To rigorously evaluate the geometric generalization capability of surrogate models, these transitional cases were excluded from training and assigned to a dedicated transitional dataset, which was only used during testing and evaluation process. Based on this flow-aware partition strategy, representative cases from both airflow regimes are selected as training samples to expose the model to characteristic flow structures while limiting the number of required training effort. A small validation dataset was additionally introduced for hyperparameter selection and early stopping during training. The remaining unseen geometries were grouped into the test dataset for final evaluation. Among the 65 generated cases, 17 cases (26.2%) are used for training, 4 cases (6.2%) for validation, and the remaining 44 cases (67.7%), including the transitional cases, were reserved for testing. Compared with random splitting, this physically informed dataset partition ensures that the surrogate models are evaluated not only on interpolation within known flow regimes but also on their capability to predict flow transitions across unseen geometries. Last but not least, prior to training, both speed and temperature fields were normalized independently into the range of [0,1] using min-max normalization to avoid scale imbalance between variables. Such normalization facilitates neural network training and improves convergence [47].

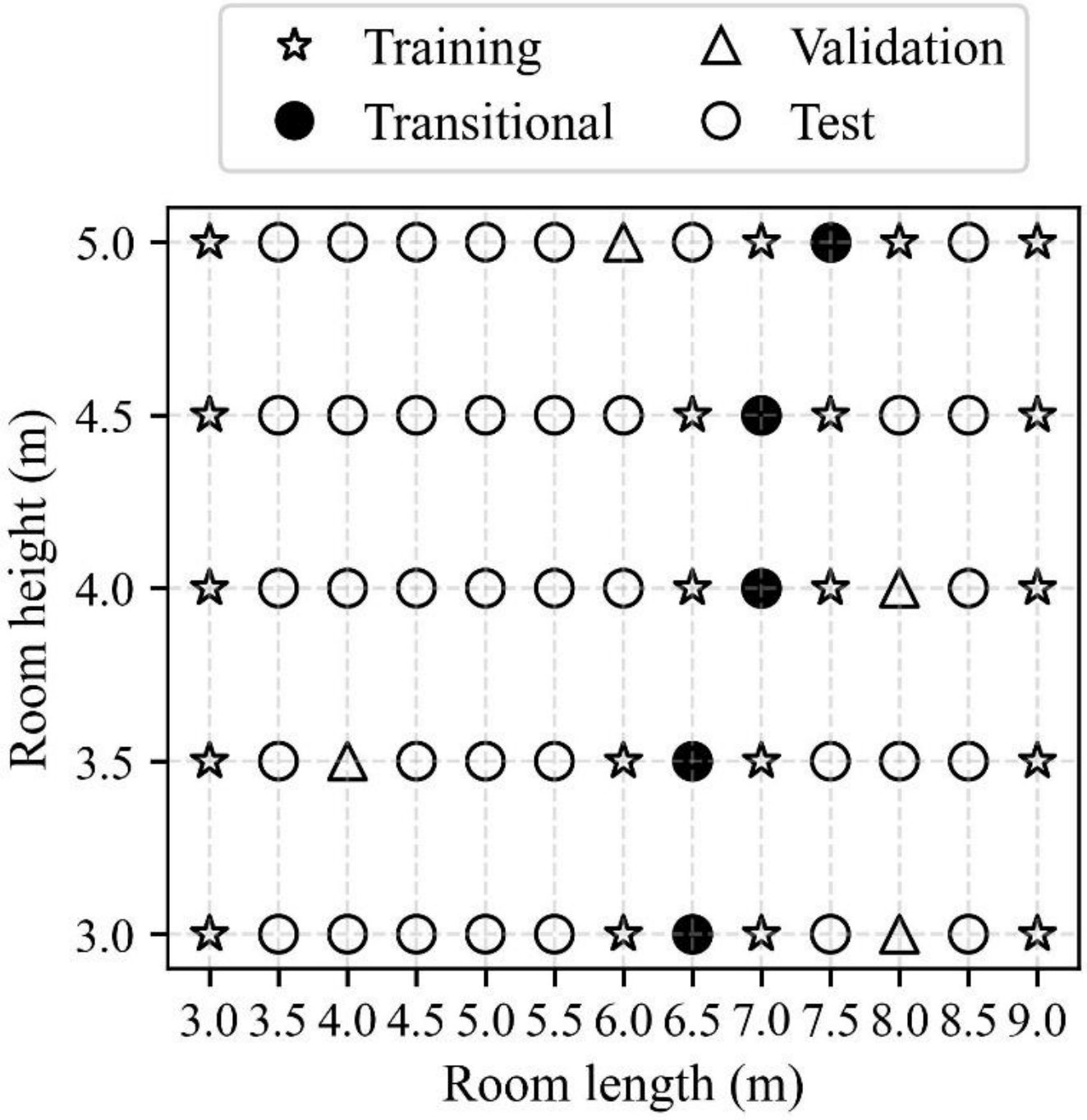

Fig. 2. Distribution of room configurations used for training, validation, and test datasets including transitional cases. Colored markers indicate representative room configurations selected for different datasets.

### 2.2 Geometry-conditioned surrogate modeling

To investigate the influence of geometry representation on surrogate modeling performance, two geometry-conditioned neural network frameworks were developed and compared. Both models aim to establish a direct mapping from room geometry to indoor airflow and temperature distributions while sharing identical training procedures and prediction targets. The overall workflow includes two steps. First, the room geometry is transformed into model inputs through dedicated geometry representation methods. Then the geometry representation is provided to the surrogate model to predict the corresponding airflow and temperature fields. To ensure a fair comparison, identical training, validation, and testing datasets are adopted for both approaches, such that the observed differences could primarily be attributed to the geometry representation strategy.

#### 2.2.1 Geometry representation

Different room geometries generate CFD fields with varying spatial dimensions, which is challenging for CNN-based learning and currently limits efficient training within a unified surrogate framework [40,41] . Therefore, we introduced geometry representations to transform rooms with different dimensions into structured inputs with consistent tensor formats while preserving essential spatial and geometric information. In this study, two representations shown in Fig. 3 were investigated: CaNN, which explicit embedding of room size variation by preserving different geometries as varying occupied regions within a fixed canvas, and CoNN, which represents room geometry through normalized coordinate fields with a fixed spatial resolution. The corresponding detailed geometry encoding procedures are illustrated in Fig. 3.

Starting from the parameterized room configurations described in Section 2.1, different room dimensions are first generated by varying room length and height while maintaining identical ventilation and thermal boundary conditions. The generated geometries are then transformed into network inputs through these two representation ways.

For CaNN, room geometry is represented explicitly through geometry embedding in a fixed-size canvas occupying different region size in the canvas. As shown in Fig. 3, all room configurations are projected into a predefined canvas with dimensions corresponding to the maximum room size with $L_{max} = 9\ m$ and $H_{max} = 5\ m$.

After embedding, each room occupied only part of the global canvas, while the remaining area is padded with zeros.

To encode the spatial position of each grid point and facilitate network training, two normalized coordinate channels $x^*$ and $y^*$ are constructed:

$$x^* = \begin{cases} \frac{x}{L_{max}}, if\ (x, y) \in \Omega \\ 0, if\ (x, y) \notin \Omega \end{cases} \tag{1}$$

$$y^* = \begin{cases} \frac{y}{H_{max}}, if\ (x, y) \in \Omega \\ 0, if\ (x, y) \notin \Omega \end{cases} \tag{2}$$

where $x$ and $y$ denote the physical coordinates in the global canvas. $\Omega$ denotes the physical room domain.

Each room with dimensions $l$ and $h$ occupies only part of the global canvas. The occupied pixel numbers are determined as:

$$n_x^{room} = round\ \left(n_x \frac{l}{L_{max}}\right) \tag{3}$$

$$n_y^{room} = round\left(n_y \frac{h}{H_{max}}\right) \tag{4}$$

Where $N_x^{room}$ and $N_y^{room}$ are occupied pixel numbers horizontal and vertical directions, respectively. $N_x$ = 126, $N_y$ = 70 represent the total numbers of pixels in the fixed canvas along the x and y directions.

In addition, a binary domain mask $M_\Omega$ is introduced to explicitly indicate the occupied computational region:

$$M_\Omega = \begin{cases} 1, (x, y) \in \Omega \\ 0, (x, y) \notin \Omega \end{cases} \tag{5}$$

Accordingly, the final input matrix $\psi_{CaNN}$ for the NN is defined as:

$$\psi_{CaNN} = (x^*, y^*, M_\Omega) \in \mathbb{R}^{3 \times 70 \times 126} \tag{6}$$

where the three channels correspond to horizontal coordinates, vertical coordinates, and domain occupancy information, respectively. By introducing a fixed global reference frame, CaNN expresses absolute geometry information including room dimensions, boundary locations, and occupied spatial extent.

After NN prediction, since the predicted fields $F_{CaNN}$ of CaNN are generated on the fixed canvas, a post-processing procedure was applied to recover the original room geometry. First, the valid room domain is retained and the region outside the room is set to zero using the domain mask:

$$F'_{CaNN} = F_{CaNN} \odot M_\Omega \tag{7}$$

Where $F_{CaNN}$ denotes the predicted field generated by CaNN on the fixed canvas, $F'_{CaNN}$ denotes the predicted field after applying the binary domain mask, $M_\Omega$ represents the binary domain mask, and $\odot$ indicates element-wise multiplication.

The extracted field is subsequently cropped to the occupied dimensions:

$$\Omega = \{(i, j) | 1 \le i \le n_x^{room}, 1 \le j \le n_y^{room}\} \tag{8}$$

$$F_\Omega = F'_{CaNN}|_\Omega \tag{9}$$

Where Ω denotes the valid room domain, $F_\Omega$ denotes cropped predicted field within the valid room domain.

Finally, the normalized coordinates are transformed back into physical coordinates through inverse normalization:

$$x = x^{*} L_{max};\ x \in [0, L] \tag{10}$$

$$y = y^{*} H_{max};\ y \in [0, H] \tag{11}$$

In contrast, CoNN adopts an implicit geometry representation strategy based solely on normalized coordinate channels. Unlike CaNN, CoNN does not employ a fixed canvas or introduce an explicit occupied domain mask. For each room geometry, two coordinate channels are directly generated over a fixed spatial resolution of 96×96. The coordinate values are normalized according to the maximum room dimensions:

$$x_i^{*} = \frac{x_i}{L_{max}}\ , \quad i=1,\ \cdot\ \cdot\ \cdot\ , 96 \tag{12}$$

$$y_j^{*} = \frac{y_i}{H_{max}}, \quad j=1,\ \cdot\ \cdot\ \cdot\ , 96 \tag{13}$$

where $L_{max}$=9 $m$ and $H_{max}$=5 $m$ . $x_i$ and $y_i$ are physical coordinates. $x_i^{*}$ and $y_j^{*}$ are normalized coordinates.

Accordingly, the coordinates of different rooms range became:

$$x \in \left[0, \frac{L}{L_{max}}\right] \tag{14}$$

$$y \in \left[0, \frac{H}{H_{max}}\right] \tag{15}$$

Finally, the normalized input tensor is defined as:

$$\psi_{CoNN} = \left[x^{*}, y^{*}\right] \in \mathbb{R}^{2\times 96\times 96} \tag{16}$$

Unlike CaNN, coordinate values are assigned to the entire input domain without masking non-occupied regions. Consequently, geometry information is encoded implicitly through the coordinate values rather than explicit domain occupancy. By this formulation, different room dimensions produce different coordinate ranges and maps while maintaining identical input dimensions.

After NN prediction, no domain extraction or cropping operation is required. The predicted field is directly mapped back to the physical coordinate system through inverse normalization:

$$x_i = x_i^* L_{max};\ x_i \in [0, L] \tag{17}$$

$$y_j = y_j^* H_{max};\ y_i \in [0, H] \tag{18}$$

Where $i, j$ denote the spatial indices of the predicted field. Thus, restoring the predicted airflow and temperature fields to the original room geometry.

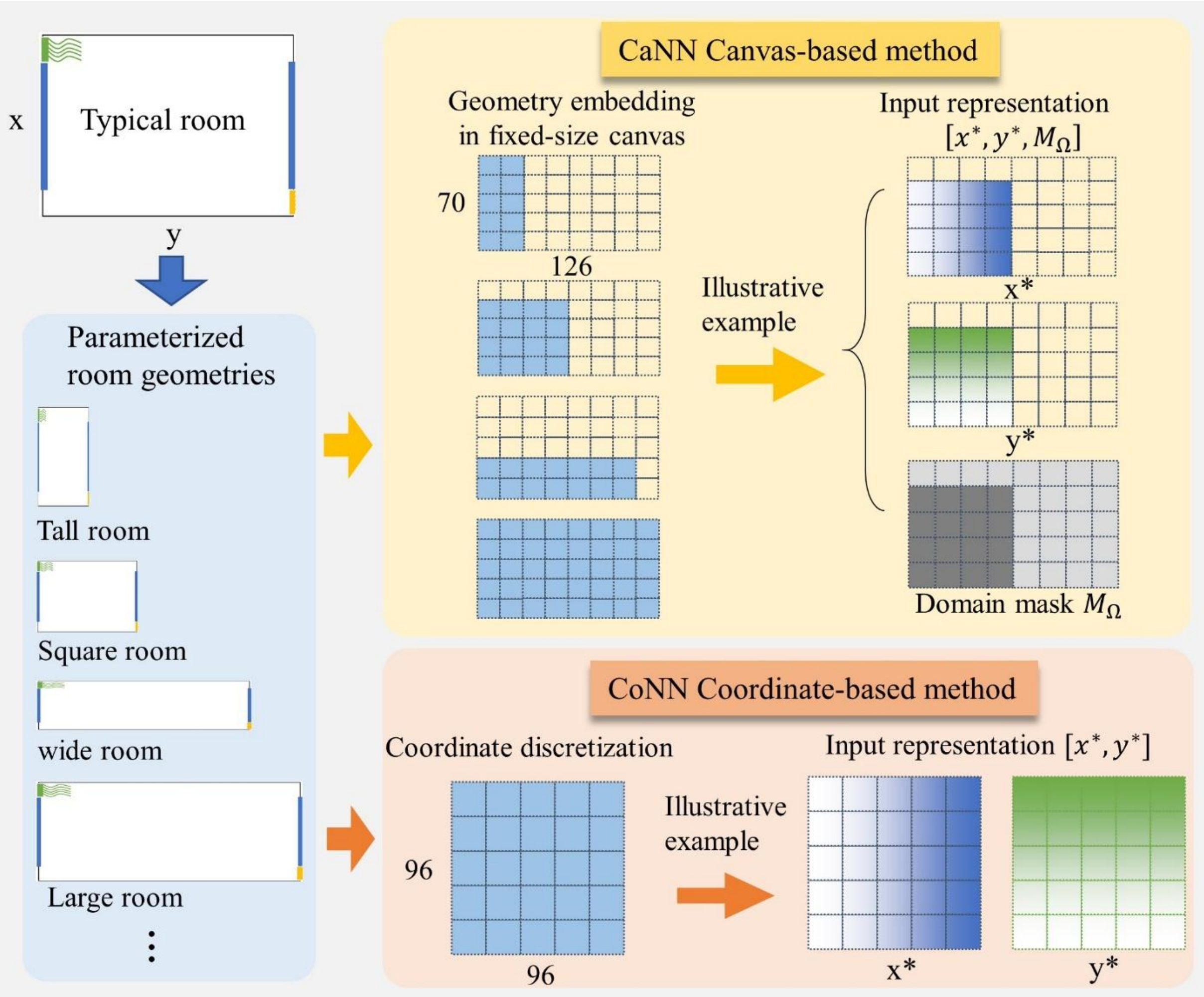

Fig. 3. Comparison between the proposed CaNN and CoNN geometry representations.

### 2.2.2 Neural network structure

After transforming room geometries into structured input representations, convolution autoencoder (CAE) surrogate model with U-Net was employed to establish the mapping between geometry information and indoor airflow fields. To isolate the influence of geometry representation, both CaNN and CoNN adopted a unified prediction framework with comparable network complexity and training settings. As illustrated in Fig. 4, the proposed network followed an encoder-decoder architecture consisting of three components: an encoder for hierarchical feature extraction, and two independent decoder branches for reconstructing the speed and temperature fields. Although the input dimensions of CaNN and CoNN differ due to the geometry encoding methods, the overall network topology, channel progression, and reconstruction strategy were maintained almost consistently to ensure that performance differences mainly originate from the geometry representation rather than network capacity.

**CaNN architecture**

The CaNN model adopts a U-Net-inspired encoder-decoder architecture with residual network (ResNet) and dual independent decoders of speed and temperature fields, as illustrated in Fig. 4. The encoder contains five encoding stages. Each stage consists of convolutional feature extraction, Rectified linear (ReLU) activation, max pooling for spatial compression, and residual refinement. The number of channels progressively increase from 3 to 64, while the spatial dimensions decrease from 70 × 126 to 3 × 4. To reconstruct multiple physical variables simultaneously, two independent decoder branches are employed for speed and temperature reconstruction. The decoder follows symmetric spatial recovery from 3 × 4 to 70 × 126 with channel changed from 64 to 3. Skip connections are applied between corresponding encoder and decoder stages through element-wise addition to preserve local geometric information. Besides, ReLU activation function is adopted in the hidden layers to introduce nonlinearity and improve training efficiency by alleviating gradient vanishing. In the output layer, a sigmoid activation function is applied to constrain the predicted values within a normalized range and facilitate stable reconstruction of the target fields. In the training process, Adam optimizer with an initial learning rate of $1 \times 10^{-4}$ and a weight decay coefficient of $1 \times 10^{-7}$ are adopted.

Last but not least, the design of loss function is critical for the model performance. Since the CaNN output is generated based on the full canvas, the padded region outside the physical room domain should not contribute to the reconstruction error. Therefore, a masked loss function is adopted.

For each predicted field, the reconstruction loss inside the valid room domain is calculated as:

$$L_{\text{in}} = \frac{\Sigma\left(F_{pred} - F_{CFD}\right)^2 M_\Omega}{\Sigma M_\Omega + \varepsilon} \tag{19}$$

where $F_{pred}$ and $F_{CFD}$ denote the predicted field and the corresponding CFD reference field, respectively, $M_\Omega$ is the binary occupancy mask, and ε is a small constant for numerical stability.

In addition, an outside-domain penalty is introduced to suppress nonphysical predictions in the padded region:

$$L_{out} = \frac{\Sigma F_{pred}^{2}\overline{M_{\Omega}}}{\Sigma(1\text{-}M_{\Omega}) + \varepsilon} \tag{20}$$

Where $\overline{M_{\Omega}}$ denotes the complementary mask corresponding to the padded region outside the valid room domain.

The loss for each physical variable is then defined as:

$$L_{overall} = L_{\text{in}} + \lambda L_{out} \tag{21}$$

where $\lambda = 0.1$ is used to balance the outside-domain penalty.

Since the model predicts both speed and temperature fields, the final CaNN loss was written as:

$$L_{CaNN} = L_{V} + L_{T} \tag{22}$$

where $L_V$ and $L_T$ denote the losses for speed and temperature prediction, respectively.

**CoNN architecture**

The CoNN model adopts the same encoder-decoder structure, residual design, dual-decoder and U-Net configuration as CaNN. Therefore, the only modifications are introduced at the input and output levels. The spatial resolution of the feature maps through the encoder is progressively reduced from 96×96 to 3×3, and subsequently reconstructed back to 96×96 through the decoder. channel progression and other configurations remain identical to that of CaNN.

The loss function is simplified to the standard mean squared error over the full field:

$$L_{CoNN} = L_{V}^{MSE} + L_{T}^{MSE} \tag{23}$$

Since CoNN does not employ canvas padding or occupancy masking, no masked loss or outside-domain regularization is applied. The prediction is directly transformed back into the physical coordinate system through inverse coordinate normalization.

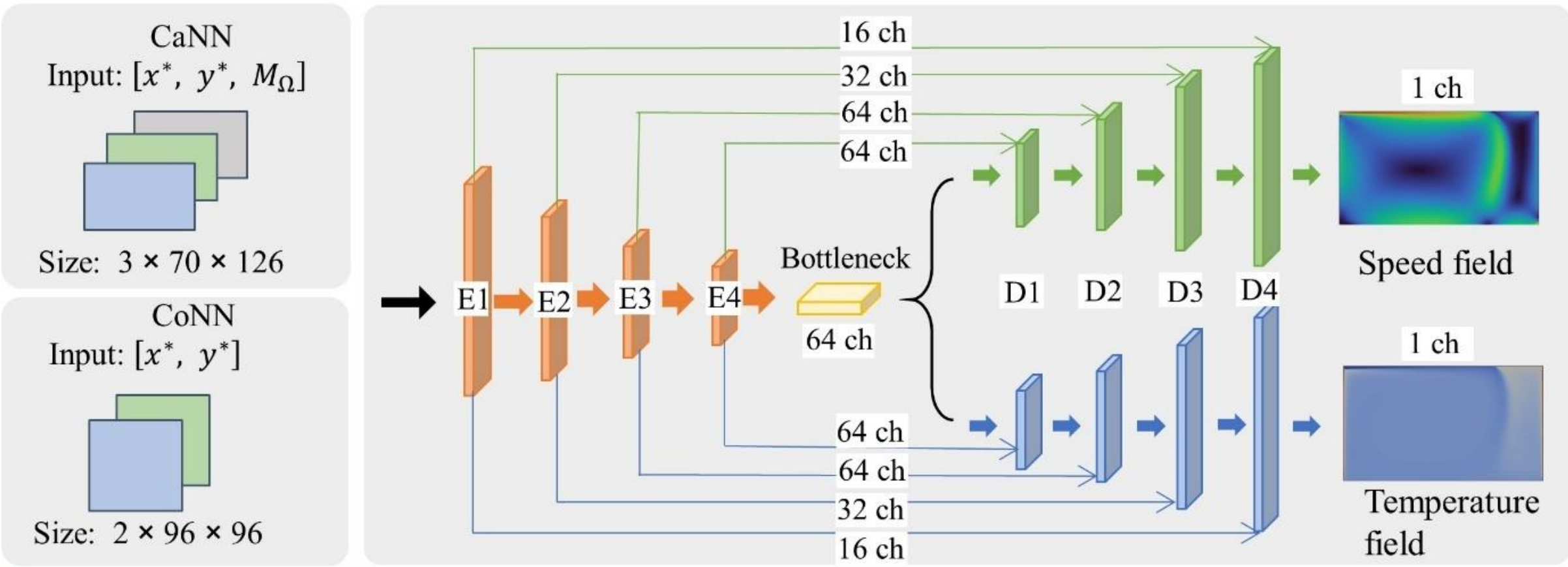

| Spatial resolution: Height × Width (E1–E4: Encoder outputs, D1–D4: Decoder outputs) | | | | | | | | | | | |
|---|---|---|---|---|---|---|---|---|---|---|---|
| Stage | Input | E1 | E2 | E3 | E4 | Neck | D1 | D2 | D3 | D4 | Output |
| Ca-NN | 70×126 | 35×63 | 18×32 | 9×16 | 5×8 | 3×4 | 5×8 | 9×16 | 18×32 | 35×63 | 70×126 |
| Co-NN | 96×96 | 48×48 | 24×24 | 12×12 | 6×6 | 3×3 | 6×6 | 12×12 | 24×24 | 48× 48 | 96×96 |

Fig. 4. Architecture of the proposed NN with dual-decoder and U-Net framework for indoor airflow and temperature field prediction using CoNN and CaNN geometry representations. The spatial resolutions of feature maps for CaNN and CoNN are summarized in the table below. E1–E4 denote the outputs of encoder stages, while D1–D4 denote the outputs of decoder stages. The number of channels at each stage is indicated above the corresponding feature maps.

## 3. Results and discussion

In this section, the indoor airflow and temperature fields predicted by CaNN and CoNN are presented, compared, and evaluated against the CFD reference results for both the training dataset and test dataset. Model performance is assessed through qualitative comparisons of field distributions as well as quantitative error metrics. Specifically, Section 3.1 presents the prediction results on the training dataset, while Section 3.2 evaluates and compares the prediction performance and generalization capability of CaNN and CoNN on the test dataset. Afterwards, beyond individual case-by-case evaluation, the overall comparison of error distribution between two models in training and test datasets is provided in Section 3.3. Last but not least, computational efficiency of the proposed approaches is analyzed and discussed.

### 3.1 Prediction performance on training dataset

The objective of evaluating the training dataset is to check whether the proposed models have successfully learned the mapping from geometry-conditioned inputs to airflow and temperature fields and are capable of accurately reconstructing the training samples. Firstly, the losses of training and validation are introduced that both models achieved stable convergence after 10,000 training epochs. Training losses decreased consistently and converged to similar values, reaching the order of $10^{-5}$, with final value of approximately $0.6 \times 10^{-5}$ for CaNN and $0.2 \times 10^{-5}$ for CoNN. The validation loss curves followed trends similar to those of the training losses and converged to comparable levels. This behavior suggests that both models successfully learned the underlying geometry-field mapping without obvious signs of underfitting or overfitting. Although the final losses of CoNN were approximately three times lower than those of CaNN, this difference alone cannot be interpreted as superior model performance, since training and validation losses only reflect optimization behavior within the available data and do not directly indicate predictive accuracy or generalization capability on test dataset.

To further evaluate the detailed learning and prediction capability of the proposed surrogate models on training dataset, four representative cases from the training set were selected for comparison, corresponding to the extreme combinations of room dimensions: (i) minimum length and minimum height (L = 3.0 m, H = 3.0 m), (ii) maximum length and minimum height (L = 9.0 m, H = 3.0 m), (iii) minimum length and maximum height (L = 3.0 m, H = 5.0 m), and (iv) maximum length and maximum height (L = 9.0 m, H = 5.0 m). Figures 5 and 6 present the predicted speed and temperature fields generated corresponding absolute error distributions.

For the airflow speed field shown in Fig. 5, both models successfully reproduce the dominant flow structures across all cases. In the compact room configuration (L = 3.0 m, H = 3.0 m), the main recirculation pattern and low-speed core region are accurately captured with negligible discrepancies. As the room length increases to L = 9.0 m, a longer airflow structure develops, accompanied by a clear transition region separating the primary circulation and downstream low-speed zone. Both CaNN and CoNN preserve the location and shape of this transition boundary with high fidelity. For taller geometries (H = 5.0 m), the vertical extension of the circulation structure is also reconstructed accurately. Even in the largest geometry (L = 9.0 m, H = 5.0 m), both models maintain good agreement with CFD predictions and reproduce the overall flow topology without visible distortion. The corresponding error maps show that reconstruction errors remain generally low throughout the domain. Slightly increased errors are mainly concentrated near regions with strong velocity gradients, particularly around the curved transition zone and recirculation boundary in the long-room configurations. These regions correspond to locations where rapid changes in momentum occur and therefore impose higher learning difficulty. Compared with CaNN, CoNN exhibits marginally smoother error distributions and reduced local fluctuations in several cases, indicating slightly more stable reconstruction of detailed flow features.

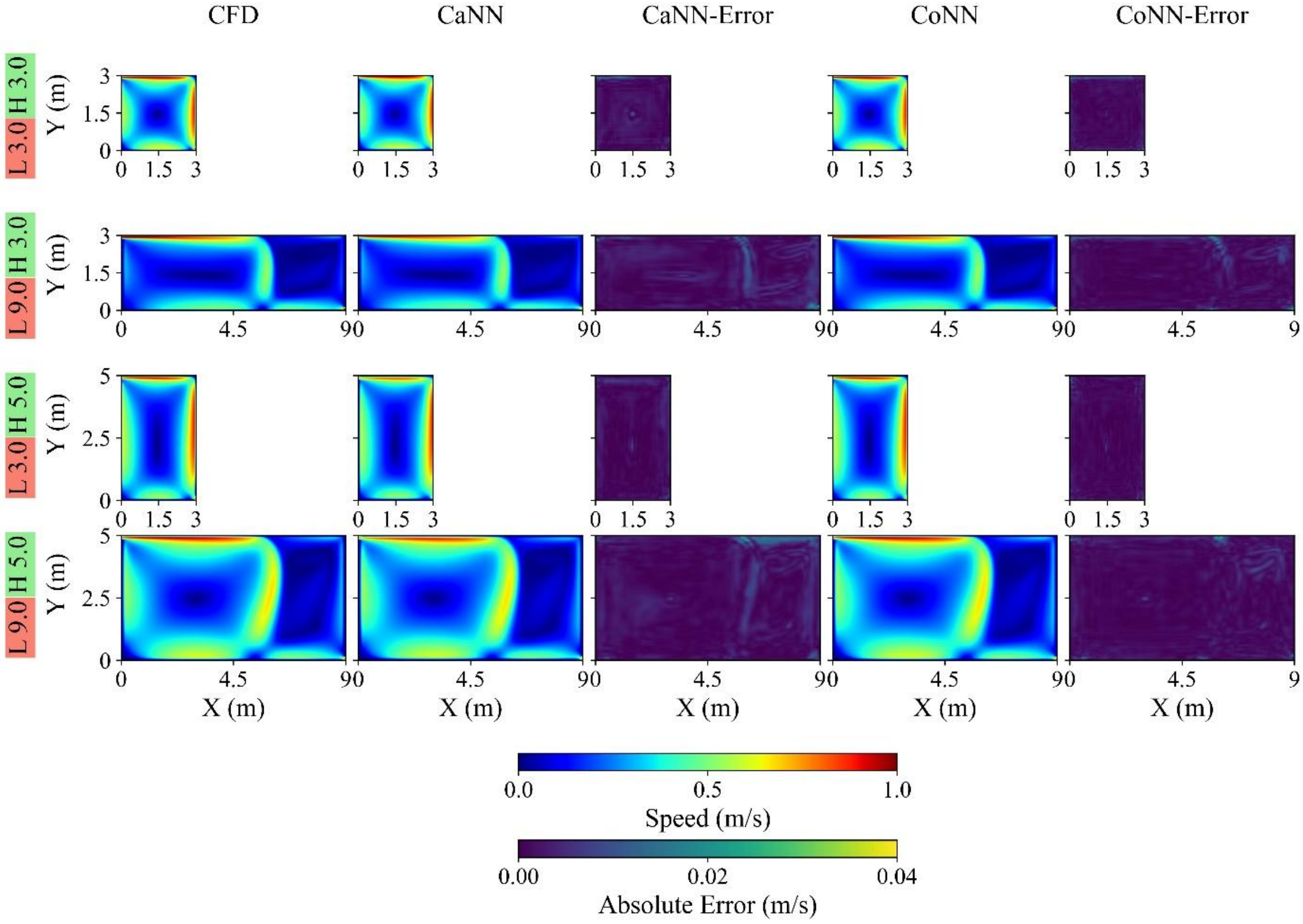

Fig. 5. Air flow speed field in the training dataset. The first column shows the CFD reference results, while the second and fourth columns present the predictions obtained by CaNN and CoNN, respectively. The third and fifth columns illustrate the corresponding absolute error distributions relative to CFD.

For the temperature field (Fig. 6), both surrogate models demonstrate similarly strong reconstruction ability. The overall thermal stratification and spatial temperature gradients predicted by both models closely match the ground truth. In short-room cases, temperature remains relatively uniform and is reconstructed almost perfectly. With increasing room length, a clear thermal transition region emerges near the downstream part of the room, which is accurately reproduced by both approaches. For taller geometries, vertical thermal distributions remain physically consistent and no noticeable problem is observed. The temperature error distributions of both models further indicate that prediction deviations are very small and spatially smooth. Besides, the errors in most of region are lower than 0.1°C in both CaNN and CoNN. Localized larger errors appear primarily around thermal interface regions where temperature gradients are strongest. For the long-room cases (L = 9 m), CaNN exhibits relatively larger prediction errors around the right window region, indicating a reduced capability to reconstruct localized temperature gradients compared with CoNN. Compared with the speed field, the temperature field exhibits visually bigger reconstruction error which is around 5 times that of speed prediction, suggesting that the smoother spatial characteristics of temperature are more difficult for both models to learn.

Overall, the results demonstrate that both CaNN and CoNN achieve high reconstruction accuracy on the training dataset, indicating that the proposed geometry-conditioned framework effectively learns the mapping between geometry representations and coupled airflow temperature distributions. Among the two geometry representations, CoNN shows a slight advantage in reconstruction of speed and temperature on training dataset.

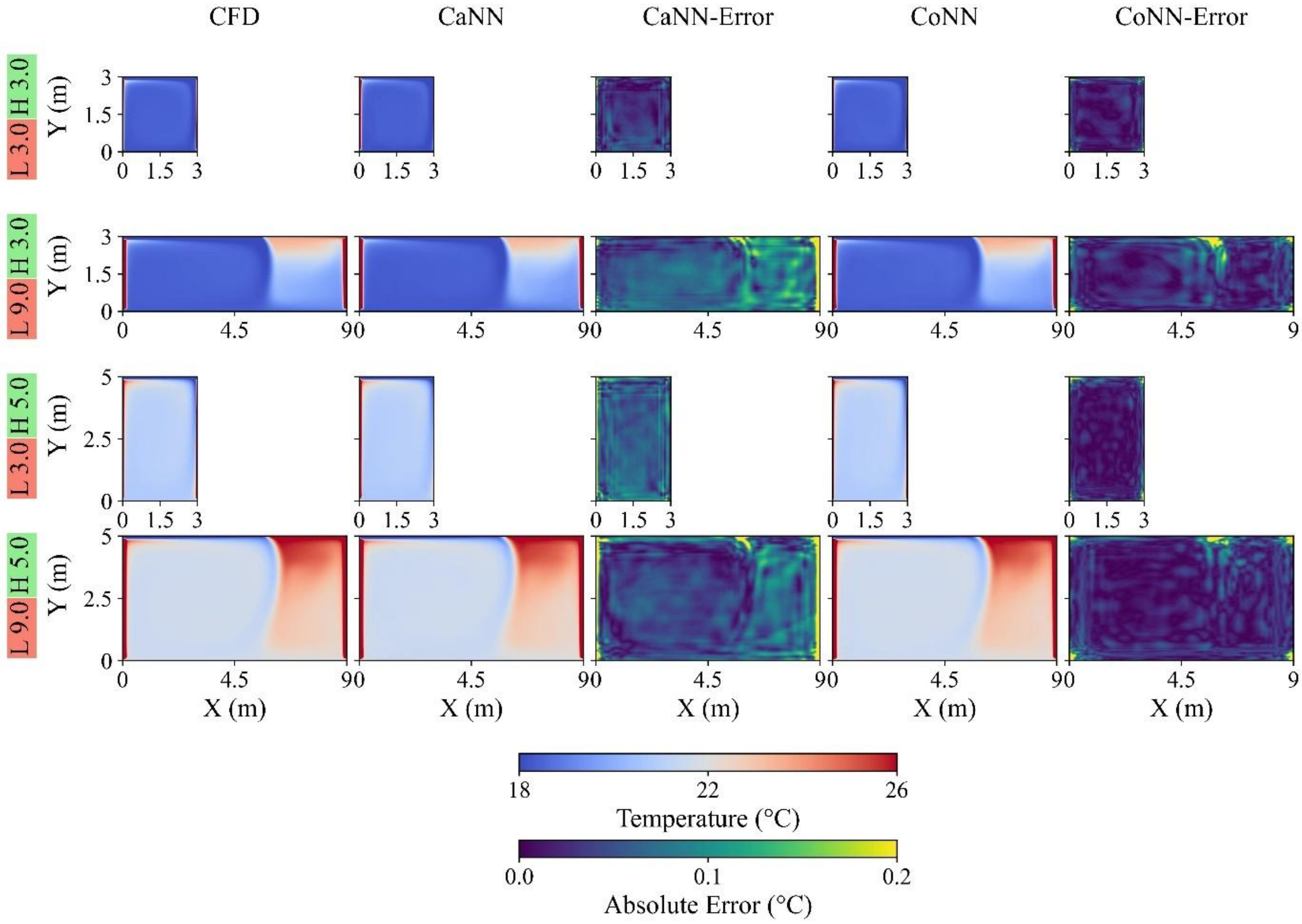

Fig. 6. Temperature field in the training dataset. The first column shows the CFD reference results, while the second and fourth columns present the predictions obtained by CaNN and CoNN, respectively. The third and fifth columns illustrate the corresponding absolute error distributions relative to CFD.

### 3.2 Prediction performance on test dataset

In this section, the prediction accuracy and geometric generalization capability of the two surrogate models are evaluated and compared using the test dataset. The results are analyzed and visualized from two perspectives. The first category investigates the influence of varying room heights and lengths. The second category focuses on the transitional cases identified for each room height, which represent the most challenging flow regimes due to the transition between single-circulation and multi-circulation structures.

### 3.2.1 Prediction under room height and length variation

Figure 7 presents the qualitative comparison of the predicted speed fields for cases with varying room heights under two representative room lengths (L = 4.0 m and L = 8.5 m). For short rooms (L = 4.0 m), both CaNN and CoNN successfully reproduced the overall airflow pattern characterized by a single dominant circulation occupying the entire room. However, noticeable differences are observed in local prediction quality. CaNN exhibited larger deviations near the right-side window region and along the circulation center, and these errors became more pronounced as room height increased from 3.0 m to 5.0 m. These deviations are likely related to the fixed-canvas representation adopted in CaNN. The sharp discontinuity between the valid room domain and the padded region may affect feature extraction near room boundaries. This effect is particularly noticeable around the heated window, where strong buoyancy-driven thermal plumes generate large velocity gradients. Consequently, small prediction offsets lead to amplified local errors. As the room height increases, the thermal plume develops over a larger vertical distance and exerts a stronger influence on the overall airflow structure, which may explain the increased deviations observed for taller rooms. In contrast, CoNN maintained closer agreement with the CFD results and better preserved the overall flow speed. For long rooms (L = 8.5 m), both models were able to capture the transition from a single circulation to separated circulation regions. Nevertheless, CaNN showed superior reconstruction of the airflow structure near the interface between the two recirculation zones, whereas CoNN generated larger local deviations around the descending jet and right part of room. The location of boundary of descending jet predicted by CoNN was displaced by around 0.3 m from the ground truth to the left, although the speed field was accurately reproduced in most regions of the room. Overall, the qualitative comparison suggests that CaNN exhibits stronger performance especially under long room conditions with complex airflow and provides more accurate reconstruction of geometry-dependent flow features.

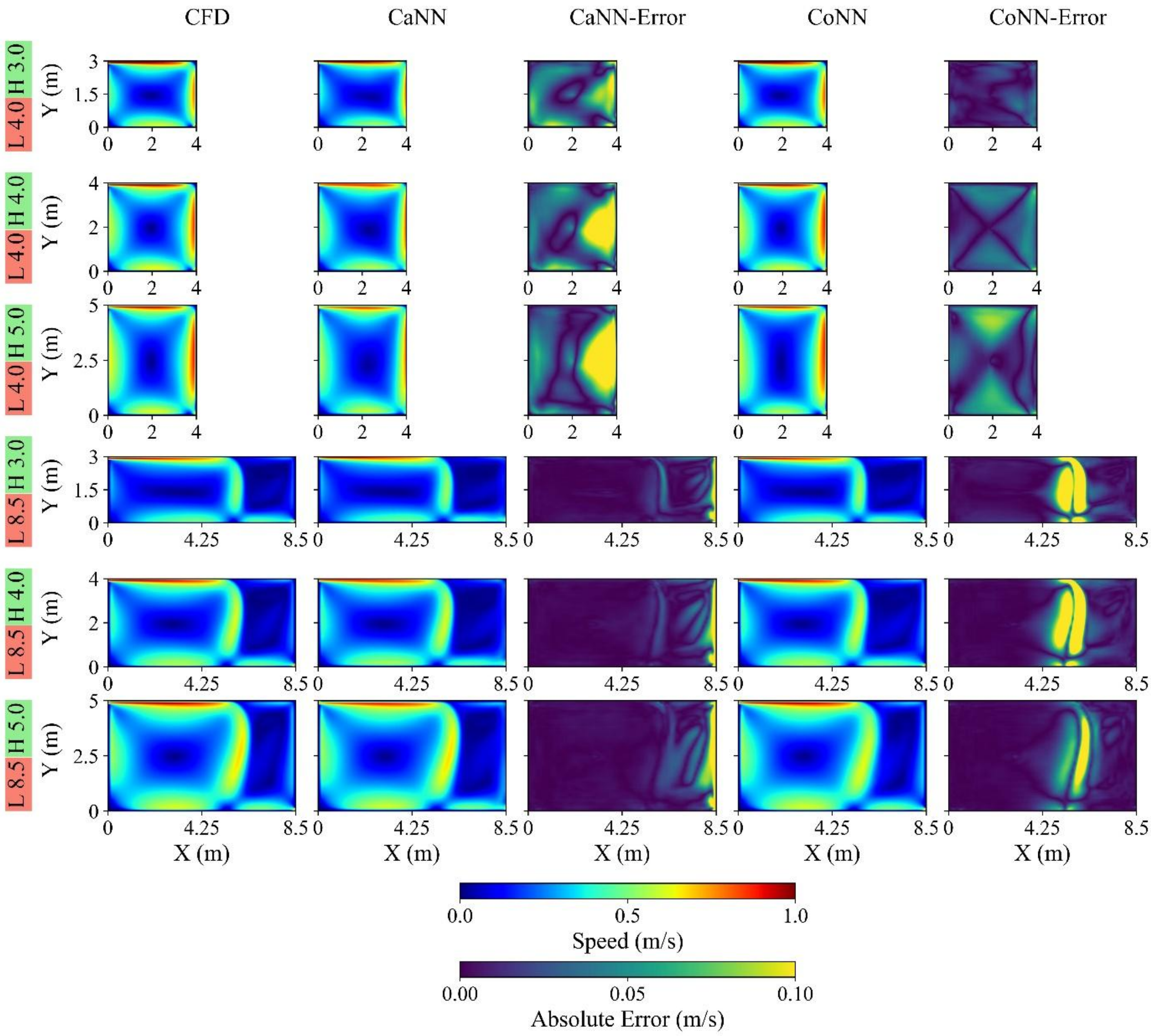

Fig. 7. Speed prediction of different room heights. The first column shows the CFD reference results, while the second and fourth columns present the predictions obtained by CaNN and CoNN, respectively. The third and fifth columns illustrate the corresponding absolute error distributions relative to CFD.

The quantitative error distributions shown in Fig. 8 further supports the visual observations. For the cases with fixed room length of L=4.0 m, CoNN consistently achieved lower absolute prediction errors than CaNN across all room heights. The median errors and interquartile ranges of CoNN remained substantially smaller, indicating both higher prediction accuracy and improved stability. Moreover, although the 95th percentile error increased with room height for both models, the growth was considerably more pronounced for CaNN, implying reduced robustness as the room geometry became taller. For the cases with L=8.5 m, both models exhibited lower overall errors than those observed for short rooms. However, CaNN has smaller error distributions and lower 95th percentile values than CoNN. Particularly, CoNN showed wider error spread and larger $95^{th}$ percentile errors of 0.16 m/s, suggesting greater sensitivity to geometric scaling under long room conditions. These results indicate that CoNN performs better for short-room cases with relatively simple airflow structures, whereas CaNN achieves superior performance in long-room cases with more complex flow conditions. This may be because CoNN relies solely on normalized coordinate representation, which is sufficient for learning simple global flow patterns, while CaNN explicitly preserves geometric occupancy information and therefore better captures effects of increasing room length.

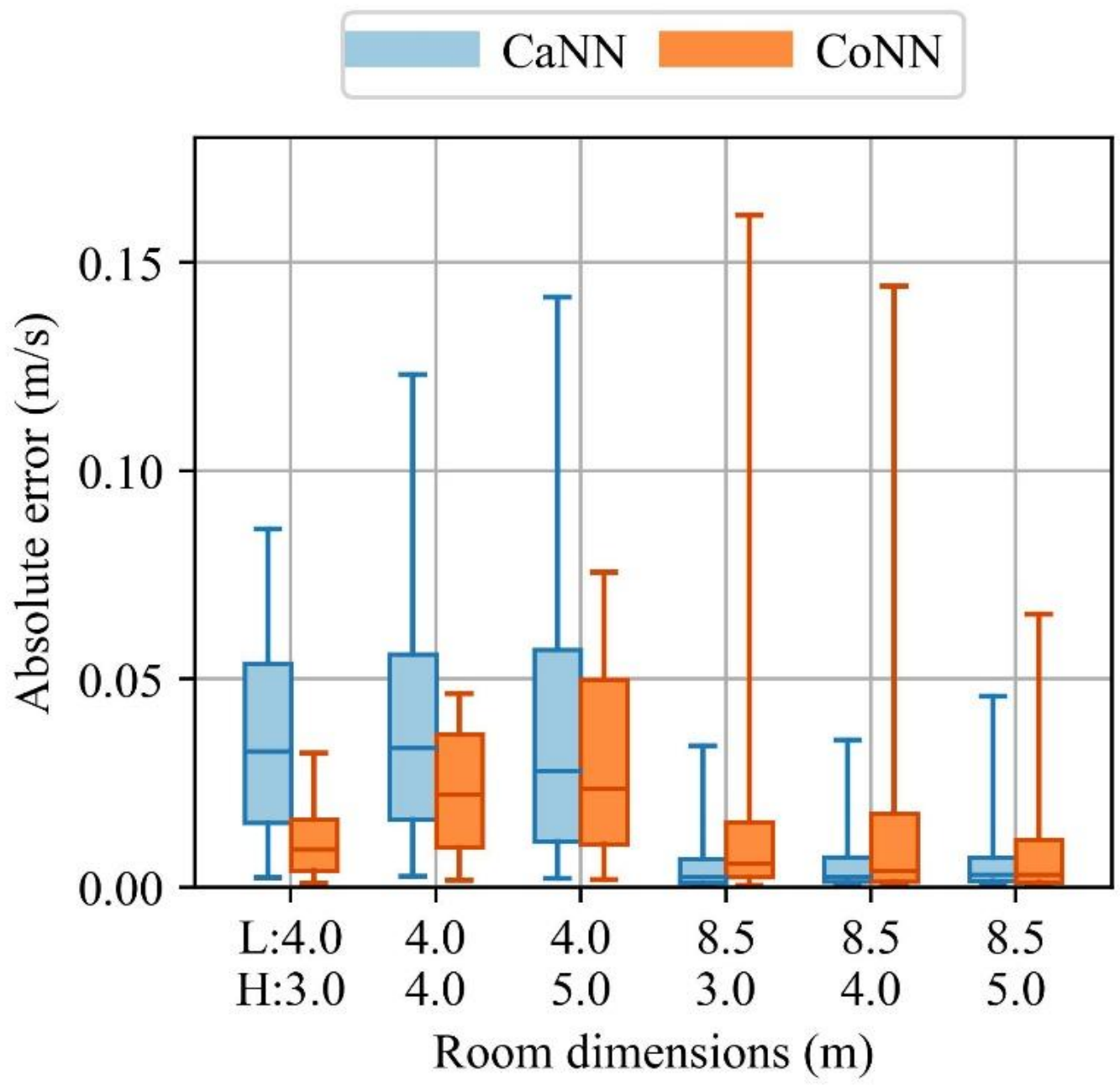

Fig. 8. Boxplot of absolute prediction errors in speed between CaNN and CoNN for representative test cases. Boxes represent the interquartile range ($25^{th}$ –$75^{th}$ percentile), and the horizontal line inside each box denotes the median value. The whiskers extend to the $5^{th}$ and $95^{th}$ percentile absolute error.

Figure 9 presents the qualitative visualization of the predicted temperature fields under varying room heights for two representative room lengths (L=4.0 m and L=8.5 m). For short-room cases (L=4.0 m), both CaNN and CoNN predicted the overall temperature distribution with relatively good agreement to CFD results. As room height increased, the thermal boundary layer near the heated window gradually developed and the temperature gradient became more pronounced. Under these conditions, CoNN generated slightly smaller local deviations and produced smoother temperature reconstruction, particularly in the central region of the room. For long-room cases (L=8.5 m), the temperature field exhibited stronger spatial non-uniformity. The left part of room is well cooled, but the right part temperature is out of control, because inlet air jet cannot reach right wall in the long room, so the left space is occupied by strong circulation of cooled air, but the right space is affected by thermal plume. This states that two surrogates show correct relation between air flow and heat transfer. Compared with CoNN, CaNN more accurately reconstructed the location and shape of the thermal transition region and produced smaller local errors around the interface between low-temperature and high-temperature zones. In contrast, CoNN showed concentrated errors near the thermal front, indicating reduced capability in resolving sharp temperature gradients under elongated room configurations. Correct prediction of transitional region is very important, because which belongs to main activity zone for people.

An interesting observation is that CaNN consistently produces larger temperature errors near the right window region. One possible explanation is that, in CaNN, room geometries are embedded into a fixed canvas, causing the heated window and associated boundary layers to appear at different spatial locations and scales across room configurations. This variation increases the difficulty of learning stable local flow and thermal patterns near the window region. In addition, the occupied room region is surrounded by zero-valued canvas areas, introducing an abrupt transition at the room boundary. Such discontinuity may further interfere with local feature extraction near the right wall. In contrast, CoNN directly encodes normalized coordinates, which preserves local positional correspondence and facilitates learning of near-wall temperature distributions.

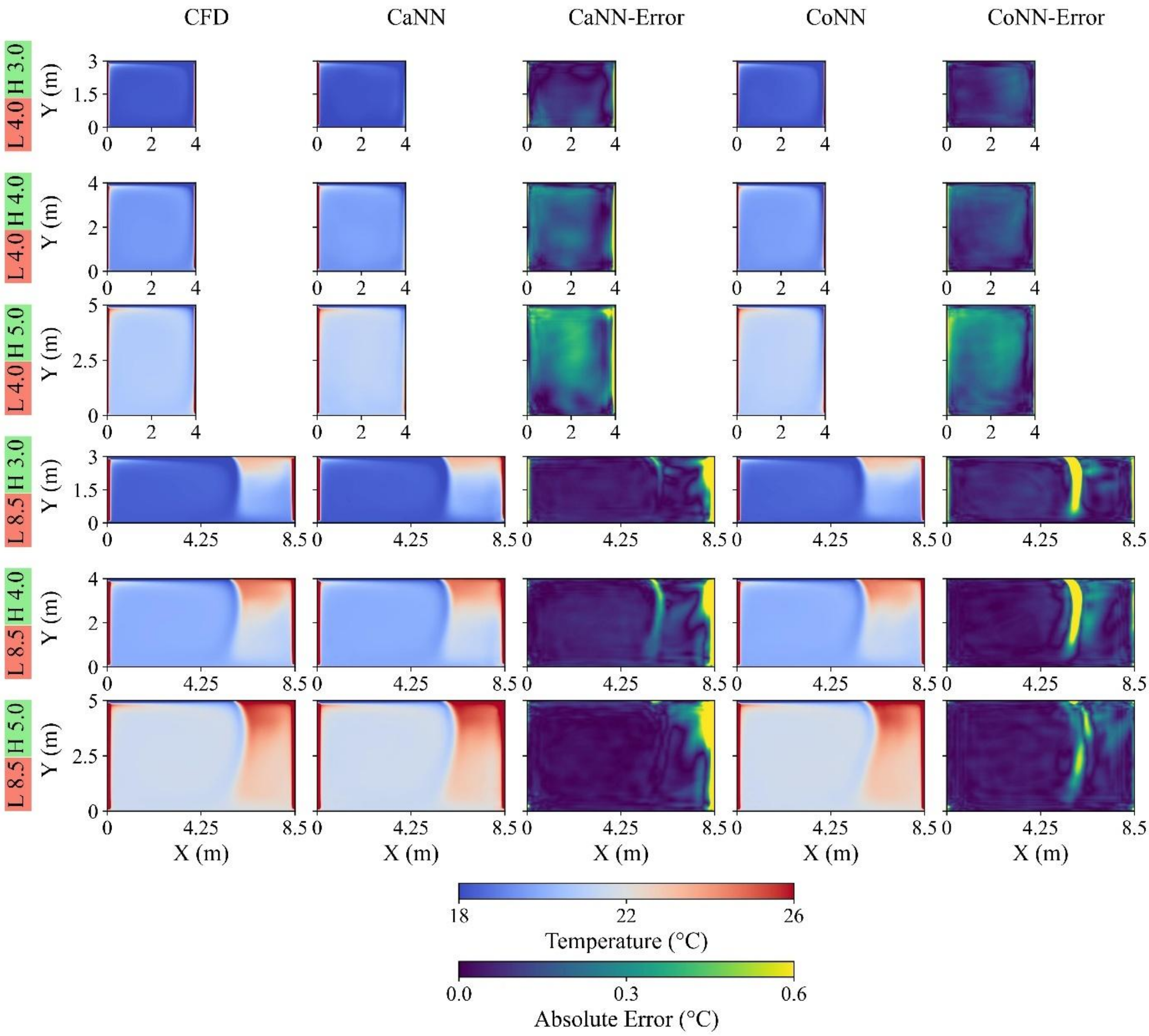

Fig. 9. Temperature prediction of different room heights. The first column shows the CFD reference results, while the second and fourth columns present the predictions obtained by CaNN and CoNN, respectively. The third and fifth columns illustrate the corresponding absolute error distributions relative to CFD.

The quantitative error distributions in Fig. 10 further confirm the observations from the temperature contours. For L = 4.0 m, both models showed increasing prediction errors with room height increase, and CoNN achieved slightly lower median errors and narrower interquartile ranges in most cases. The maximum 95th percentile error of CoNN is around 0.4 °C, which is only 0.5 °C lower than CaNN. This indicates both models are effective for reconstructing relatively smooth temperature distributions in compact rooms. For L = 8.5 m, however, CaNN generally achieved lower median errors and smaller 95th percentile values than CoNN. These results indicate that CoNN performs better for short-room cases with relatively simple thermal distributions, whereas CaNN achieves a little better performance in long-room cases with stronger temperature gradient than CoNN. This may be because the explicit geometry occupancy information in CaNN better preserves thermal structures and geometric constraints, leading to improved reconstruction of temperature gradients under complex spatial conditions.

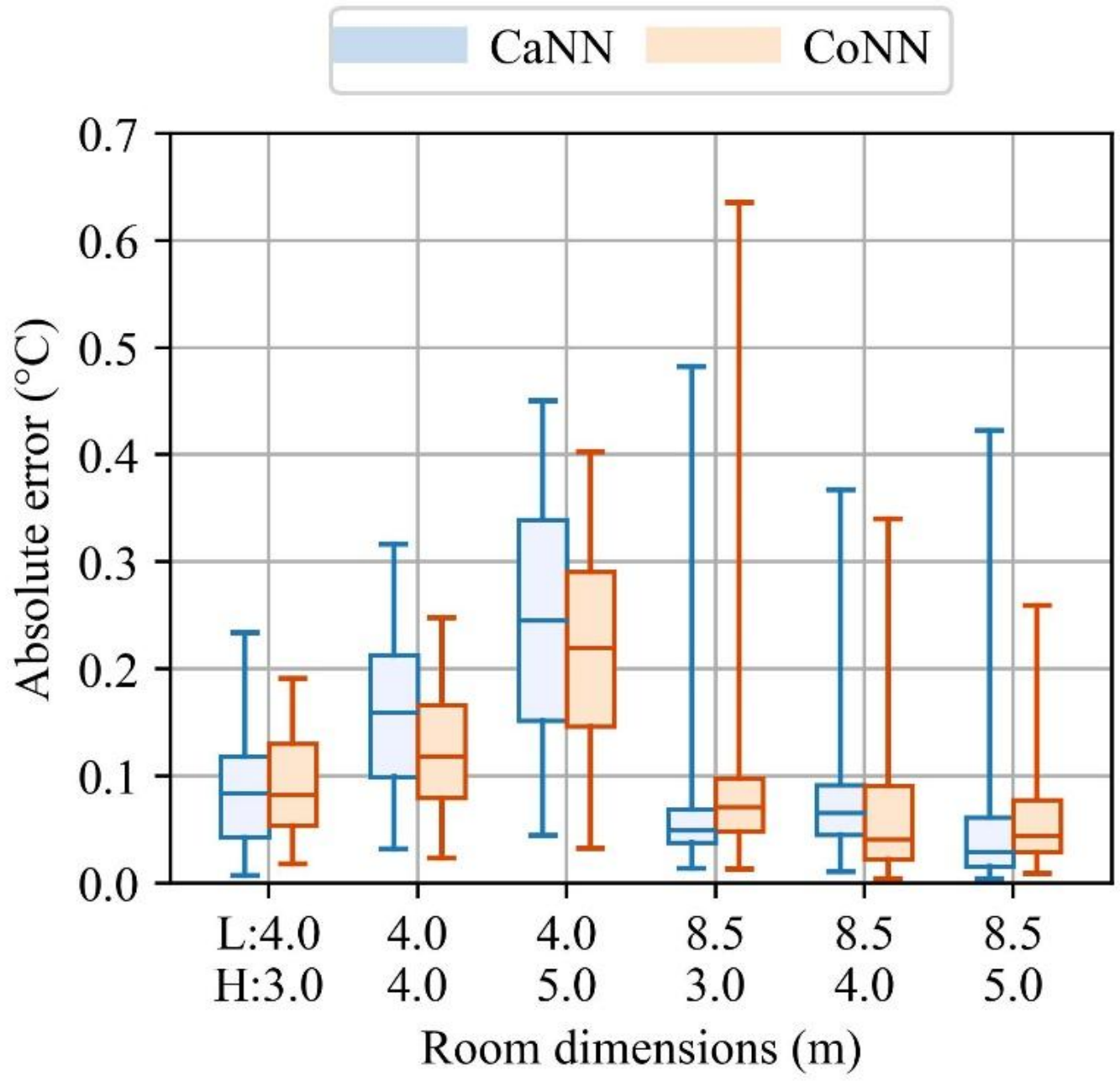

Fig. 10. Boxplot of absolute prediction errors in temperature between CaNN and CoNN for representative test cases with various room heights. Boxes represent the interquartile range ($25^{th}$ –$75^{th}$ percentile), and the horizontal line inside each box denotes the median value. The whiskers extend to the $5^{th}$ and $95^{th}$ percentile absolute error.

### 3.2.2 Prediction of transitional flow regime

Fig. 11 presents a representative transition case (H = 4.5 m, L = 7.0 m), where the indoor airflow evolves from a single-circulation regime to a double-circulation regime as the room length increases. The CFD solution indicates the emergence of a secondary recirculation region, marking a transition in the global flow topology. While CaNN successfully captures the evolving flow structure, CoNN fails to predict the formation of the secondary circulation and instead retains a flow pattern characteristic of the single-circulation regime. The white isoline (0.2 m/s) in the speed field shows that the descending jet predicted by CaNN is located at approximately 1.5 m to the right wall which is similar to the field of CFD. However, the descending jet predicted by CoNN is patched on the right wall, ignoring the effect of increasing the room length. This huge disparity is unacceptable because it completely predicts the wrong indoor flow field.

The impact of this flow-structure error is further reflected in the temperature field. In the CFD ground truth, the interaction between the recirculation zones and buoyancy effects generates a localized warm region and a distinct thermal plume near the right wall. The white isoline (22°C) in temperature field shows that the shape cooled area in CaNN is similar to that of CFD. In contrast, CoNN predicts a much larger cold air region extending across the room and fails to reproduce the thermal plume structure. Consequently, the thermal stratification and the extent of the warm region are significantly underestimated.

These results demonstrate that the inability of CoNN to capture the flow-pattern transition leads not only to errors in the velocity field but also to an incorrect representation of the underlying heat transport mechanisms. The prediction error therefore originates from a misidentification of the flow regime rather than from local reconstruction inaccuracies.

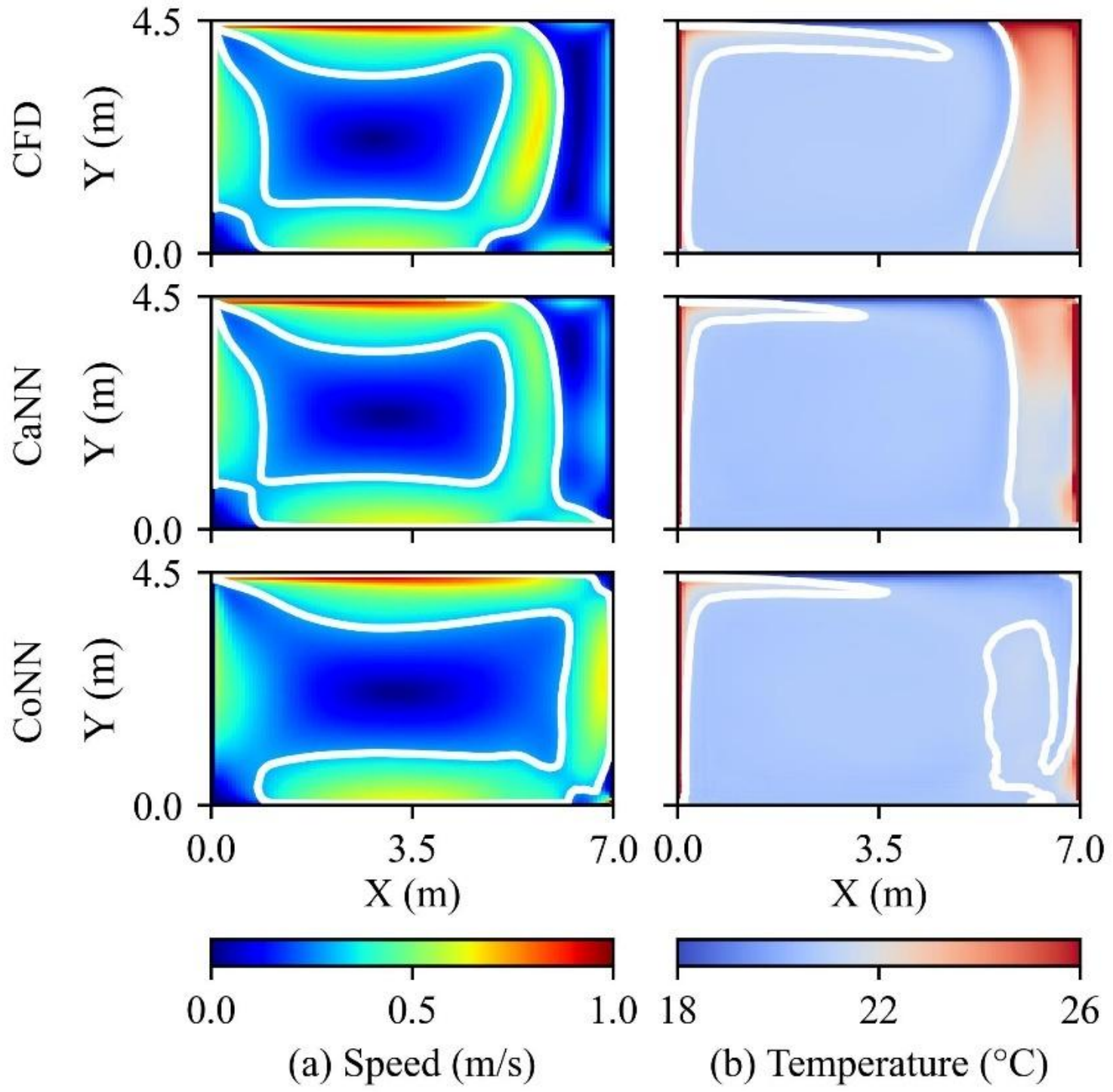

Fig. 11. Speed and temperature prediction on patter-transitional case (L = 7.0 m and H = 4.5 m). White isolines (0.2 m/s and 22°C) highlight the boundary of airflow circulation and thermal patterns.

Fig. 12 compares distribution of the 95th-percentile speed prediction error of CaNN and CoNN across the room geometry parameter space. Firstly, the training dataset shows very low prediction error around 0.02 m/s marked by five-pointed stars, indicating that both models accurately reproduce the flow fields included in the training dataset. The testing dataset presented by circle signal is divided into two parts. The first part consists of normal cases with stable flow pattern, marked by hollow circle. The second category comprises special case with transitional flow patten. For most shorter rooms, CoNN achieves prediction errors comparable to or lower than those of CaNN, demonstrating its ability to accurately reconstruct the dominant airflow structures under relative simple flow conditions. However, a different trend is observed for longer rooms, where CaNN consistently outperforms CoNN and exhibits lower prediction errors.

The prediction errors of transitional cases in testing dataset marked by solid-circles are very noticeable. These cases correspond to the critical point that the flow pattern transfer from the single-circulation to double-circulation flow regimes. The prediction errors of CaNN are bigger than other normal condition cases, but remain relatively stable. In contrast, the prediction errors of CoNN exhibits a pronounced increase reaching the range of 0.25 and 0.37 m/s, which are overall around three times bigger than that of CaNN.

The contrasting behavior of the two models suggests that CoNN is more sensitive to flow-pattern transitions. In these cases, small changes in room geometry induce a qualitative change in the global airflow topology, leading to the formation of additional recirculation zones. The results indicate that CoNN tends to preserve the dominant flow pattern learned from neighboring geometries and therefore struggles to identify the correct flow regime near the transition boundary. In contrast, CaNN demonstrates better robustness in capturing the pattern changes, resulting in lower prediction errors for the transitional cases.

These findings highlight that the main limitation of CoNN is not its performance in stable flow regime prediction, but its reduced capability to predict flow-pattern transitions. Consequently, transitional cases represent the most challenging scenarios for the model and dominate the overall prediction error distribution.

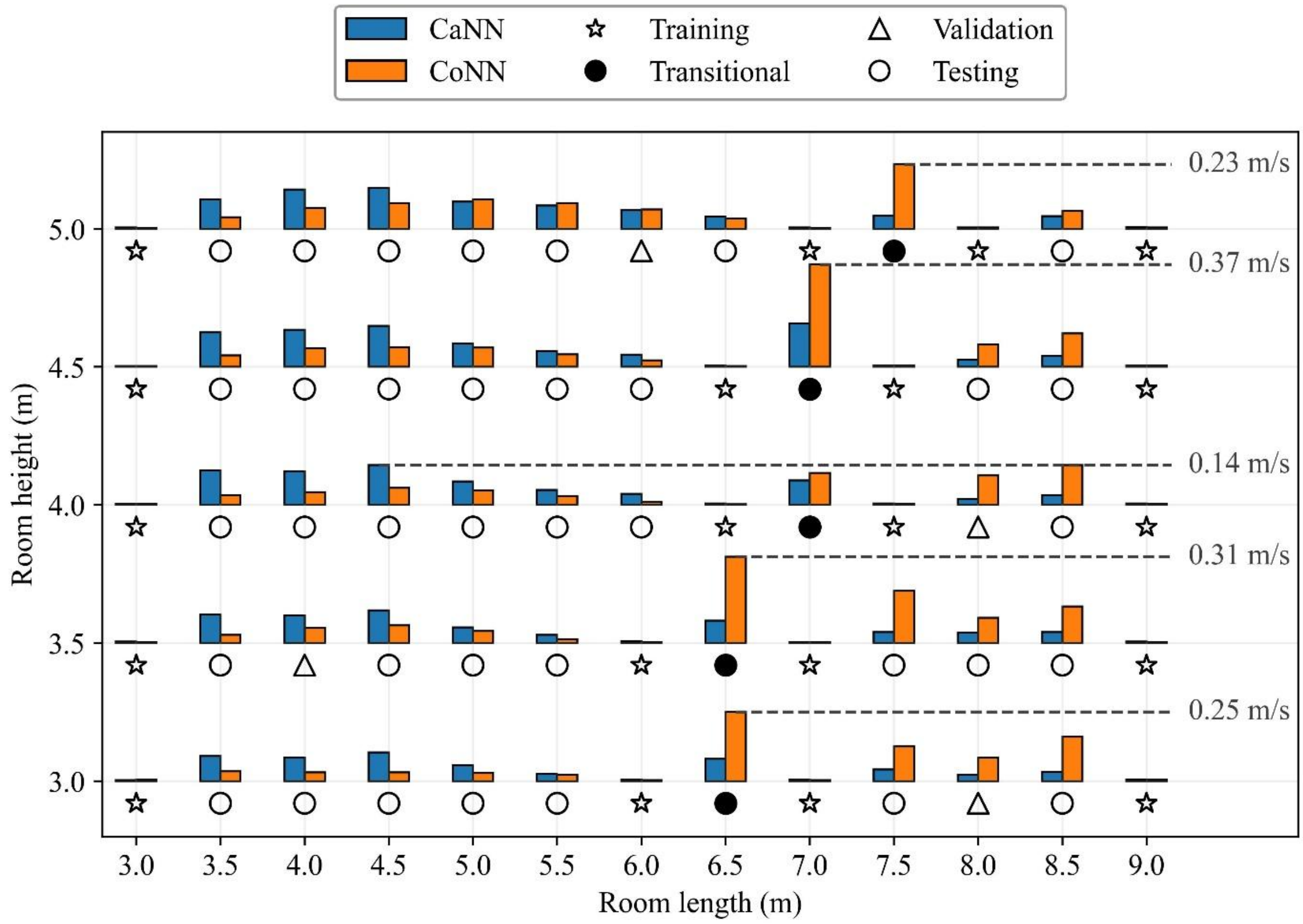

Fig. 12. Distribution of the 95th-percentile speed prediction error over the room geometry parameter space for CaNN and CoNN. Training, validation, transitional, and testing cases are indicated by stars, triangles, solid circles and hollow circles, respectively. The maximum error at each room height is highlighted.

### 3.3 Global statistical evaluation

After evaluating representative cases individually in Sections 3.1 and 3.2, a global statistical analysis is conducted to quantitatively assess the overall prediction performance of the surrogate models across the complete test dataset. While case-by-case comparisons provide detailed insights into local flow structures and thermal distributions, they may not fully reflect the overall prediction consistency and generalization capability. Therefore, Fig. 13 presents the statistical distribution of prediction errors for CaNN and CoNN on both the training and test datasets. The histograms and corresponding curves describe the percentage of samples falling within different error intervals, while the dashed vertical lines indicate the error thresholds.

For the training dataset, CoNN achieves lower prediction errors than CaNN for both speed and temperature fields, as shown in Fig. 13 (a) and (b). The speed error distribution of CoNN is more concentrated in the low-error region and exhibits a lower 95th percentile error, which is approximately 50% lower than that of CaNN, indicating stronger fitting capability for the speed field on the training samples. For temperature prediction, approximately 50% of the prediction errors are smaller than 0.03 °C, and 95% of the errors remain below 0.1 °C for CoNN. These values are noticeably lower than those of CaNN, indicating superior fitting performance for the temperature field as well. This suggests that the coordinate-based representation enables CoNN to learn the geometry-field relationship more precisely under geometrical conditions included in the training dataset.

However, the trend is reversed on the test dataset. As shown in Fig. 13 (c) and (d), CaNN exhibits lower 95th percentile errors, around 0.10 m/s and 0.58 °C for speed and temperature prediction, respectively. In contrast, CoNN shows higher 95th percentile errors of 0.15 m/s and 0.78 °C, respectively. Moreover, CaNN presents a more concentrated error distribution for both speed and temperature fields, indicating more stable predictive performance under unseen geometrical configurations. Compared with CaNN, CoNN experiences a more pronounced performance degradation from the training dataset to the test dataset, suggesting weaker generalization capability. This indicates that CaNN maintains more stable prediction performance and is capable to unseen geometrical variations, whereas CoNN tends to rely more strongly on patterns learned from the training data showing relative overfitting behavior. These results demonstrate that lower training errors do not necessarily translate into better predictive performance under unseen conditions.

The contrast between training and testing performance reveals a trade-off between fitting ability and generalization capability. Although CoNN achieves superior fitting performance on the training data, it appears to rely more strongly on the geometrical patterns observed during training. By comparison, the canvas-based representation adopted in CaNN enables more robust transfer across different geometrical conditions, resulting in improved generalization performance. Overall, the global statistical evaluation confirms that CoNN performs better in fitting a model to the training samples, whereas CaNN demonstrates superior robustness and generalization capability when predicting unseen indoor geometries.

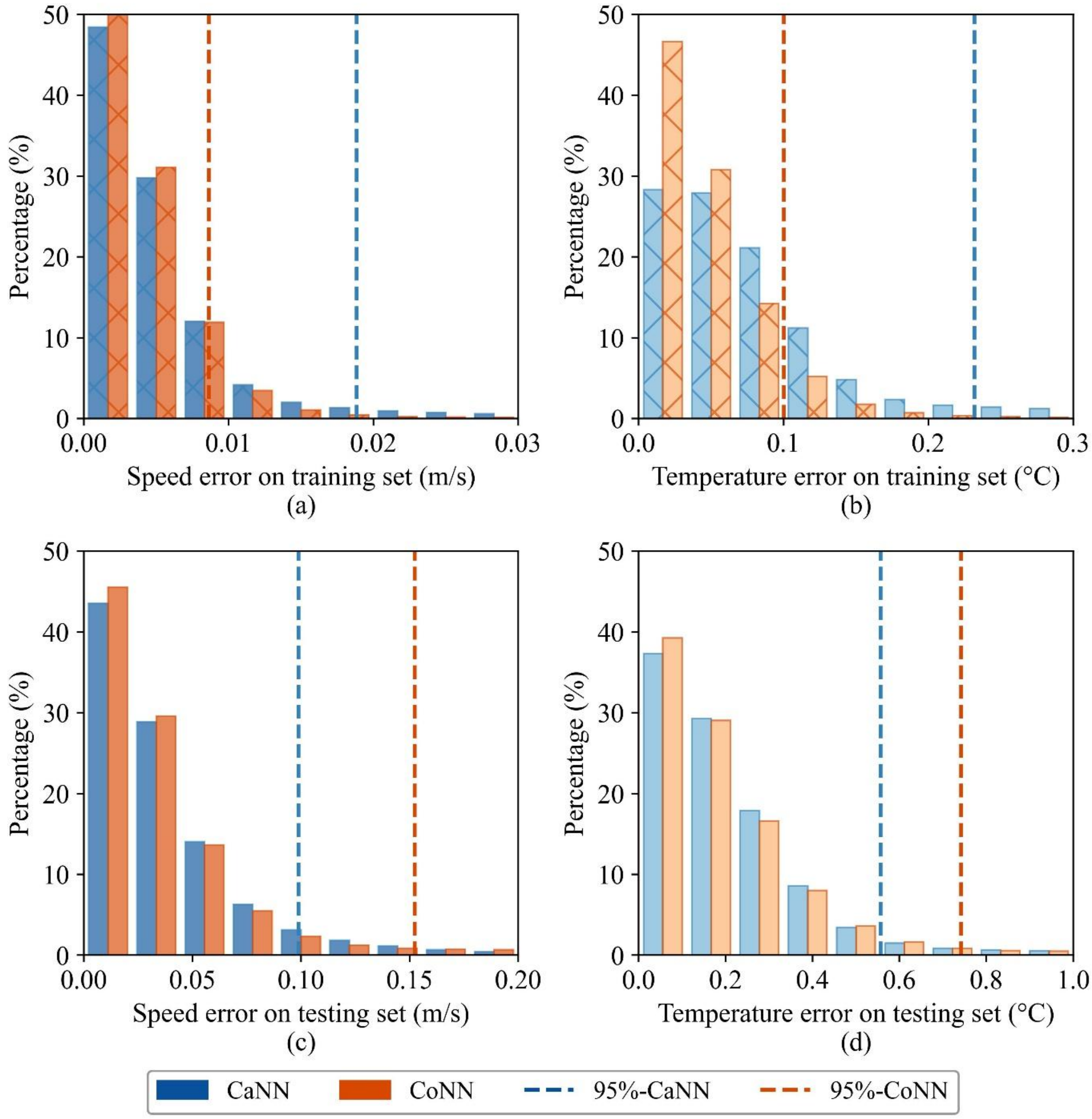

Fig. 13. Distribution of prediction errors for CaNN and CoNN on the training and test datasets: (a) speed prediction error on the training set, (b) temperature prediction error on the training set, (c) speed prediction error on the test set, and (d) temperature prediction error on the test set. The histograms and corresponding curves show the percentage distribution of prediction errors, while the dashed vertical lines indicate the error thresholds.

### 3.3.3 Computational efficiency

The computational cost of the proposed surrogate models was compared with that of conventional CFD simulations. For a typical room configuration, CFD required approximately 5 - 10 min to obtain converged velocity and temperature fields, whereas CaNN and CoNN generated predictions within around 0.1 s. Although the surrogate models require a training stage, the prediction time is reduced by several orders of magnitude. In addition, the preprocess of CFD simulation including geometry and mesh generation also require about 5 - 10 min per case. Moreover, as room size increases, the time of mesh generation and CFD computation are also enlarged. In contrast, prediction time of surrogate models remains stable and is insensitive to the room size. Although, the relative ease of solving an individual CFD case does not eliminate the practical value of surrogate modeling for repeated evaluations across many room dimensions. This computational advantage makes the proposed framework suitable for massive and rapid prediction of design exploration and real-time assessment of indoor environmental performance.

## 4. Conclusion

The developed methods CaNN and CoNN based on a U-Net architecture are able to capture and predict indoor air flows and temperature across various room lengths and heights. This paper systematically compared the prediction performance of CaNN and CoNN qualitative visualization of speed and temperature fields as well as quantitative error analysis. The main findings are summarized as follows:

(1) For short-room cases with only one circulation occupying whole room, the prediction performance of CoNN is better than CaNN. However, this trend was reversed for long rooms with more complex flow structures.

(2) For cases under transitional flow regimes, which are critical for prediction quality, CaNN successfully captured the critical room length at which the supply air jet could no longer reach the right wall and the flow pattern began to transition. In contrast, CoNN failed in 3 out of 5 transitional cases, overestimating inlet air velocity and momentum and consequently predicting airflow attachment to the right wall, which consequentially overestimated the cooling effect leading to lower indoor temperature.

(3) CaNN showed overall better agreement with CFD ground truth than CoNN, achieving 95% of the predictions fell within absolute errors of 0.1 m/s and 0.58 °C for speed and temperature prediction, respectively. These results demonstrate the potential of the proposed framework as a fast surrogate model for accurately predicting airflow and thermal field across a wide range of room geometries during the early stages of building design, thereby reducing the reliance on computationally expensive CFD simulations.

(4) Although CoNN in some cases achieved lower training errors, CaNN exhibits superior generalization capability under unseen geometrical conditions in test dataset that were associated with airflow pattern transitions where the circulation structure changes from a single-circulation regime to a double-circulation regime. This indicates that lower fitting errors do not necessarily translate into better predictive performance.

(5) The geometry representation strategy significantly influences surrogate generalization performance. Compared with CoNN which only using normalized coordinate channels, CaNN provides a more explicit embedding of room size variation by preserving different geometries as varying occupied regions within a fixed canvas, enabling the network to more effectively perceive geometric scaling and improving prediction robustness under complex and transitional flow conditions.

(6) Both CaNN and CoNN substantially improved prediction efficiency compared with conventional CFD simulations and achieved 0.1s prediction of speed and temperature field per case. This prediction time enable rapid evaluation of indoor airflow and thermal distributions in rooms with varying dimensions, supporting design-stage assessment of whether conditioned supply air effectively provide fresh and cool air throughout the occupied space.

Overall, CaNN and CoNN both are capable of prediction on most of cases across different room dimensions, but CaNN outperforms CoNN in prediction accuracy and $95^{th}$-percentile error, and the reconstruction of circulation and thermal patterns, particularly for rooms near the flow-pattern transition boundary. Future work will extend proposed framework beyond variations in room dimensions to investigate its generalization capability for more diverse and complex room geometries. In addition, more advanced geometry representations will also be explored for better prediction performance.

## 5. Acknowledgement

The author gratefully acknowledges support from the Europa-Programme (Grant No. 2014/C 198/01) and the DFG Heisenberg Programme (Grant No. GE 1652/4-1).